\documentclass[runningheads]{llncs}
\usepackage{xspace}
\makeatletter
\DeclareRobustCommand\onedot{\futurelet\@let@token\@onedot}
\def\@onedot{\ifx\@let@token.\else.\null\fi\xspace}
\def\eg{\emph{e.g}\onedot} 
\def\ie{\emph{i.e}\onedot} 
 
\def\etc{\emph{etc}\onedot} \def\vs{\emph{vs}\onedot}
 
\def\etal{\emph{et al}\onedot}
\makeatother
\renewcommand{\thefootnote}{\fnsymbol{footnote}}
\usepackage{graphicx}
\usepackage{comment}
\usepackage{cite}
\usepackage{epsfig}
\usepackage{algpseudocode}
\usepackage{algorithm}
\usepackage{dsfont}
\usepackage[export]{adjustbox}
\usepackage{amsmath,amssymb} %
\usepackage{multirow}
\usepackage{bbm}
\usepackage{color}
\newcommand{\centered}[1]{\begin{tabular}{l} #1 \end{tabular}}
\usepackage{capt-of}%
\usepackage{booktabs}
\usepackage{varwidth}
\newsavebox\tmpbox
\definecolor{pred}{rgb}{0.0,0.28,0.73} %
\definecolor{gt}{rgb}{0.0,0.42,0.31}
\newcommand{\pred}[2]{\color{pred}\scriptsize{ #1\%/#2\%} \color{black}}
\newcommand{\gt}[2]{\color{gt}\tiny (#1\%/#2\%) \color{black}}
\usepackage[normalem]{ulem} %

\usepackage[table]{xcolor}
\definecolor{lightblue}{RGB}{225, 225, 255}
\definecolor{lightred}{RGB}{255, 225, 225}

\usepackage[pagebackref=true,breaklinks=true,colorlinks,bookmarks=false]{hyperref}

\renewcommand{\paragraph}[1]{\vspace{3pt}\noindent\textbf{#1 }}

\begin{document}
\pagestyle{headings}
\mainmatter
\def\ECCVSubNumber{4294}  %

\title{Predicting Visual Overlap of Images Through Interpretable Non-Metric Box Embeddings} %

\titlerunning{Predicting Visual Overlap of Images}
\author{
Anita~Rau\textsuperscript{$1^\star$}
\and
\hspace{4pt}
Guillermo~Garcia-Hernando\textsuperscript{$2$}
\and
\hspace{4pt}
Danail~Stoyanov\textsuperscript{$1$}
\and
\\
\vspace{2pt}
Gabriel~J.~Brostow\textsuperscript{$1,2$} 
\and
\hspace{2pt}
Daniyar~Turmukhambetov\textsuperscript{$2$}
} 
\index{Rau, Anita}
\index{Garcia-Hernando, Guillermo}
\index{Stoyanov, Danail}
\index{Brostow, Gabriel J.}
\index{Turmukhambetov, Daniyar}

\authorrunning{A. Rau et al.}
\institute{
\textsuperscript{$1$}University College London \qquad  \textsuperscript{$2$}Niantic\\
\vspace{3pt}
\url{www.github.com/nianticlabs/image-box-overlap}}
\maketitle
\footnotetext[1]{Work done during an internship at Niantic.}
    
\begin{abstract}
To what extent are two images picturing the same $3$D surfaces? Even when this is a known scene, the answer typically requires an expensive search across scale space, with  matching and geometric verification of large sets of local features. This expense is further multiplied when a query image is evaluated against a gallery, \eg in visual relocalization. While we don't obviate the need for geometric verification, we propose an interpretable image-embedding that cuts the search in scale space to essentially a lookup.

Our approach measures the \textbf{asymmetric} relation between two images. The model then learns a scene-specific measure of similarity, from training examples with known $3$D visible-surface overlaps. The result is that we can quickly identify, for example, which test image is a close-up version of another, and by what scale factor. Subsequently, local features need only be detected at that scale. We validate our scene-specific model by showing how this embedding yields competitive image-matching results, while being simpler, faster, and also interpretable by humans. 

\end{abstract}

\section{Introduction}
\begin{figure}[!ht]
    \centering
    \includegraphics[width=\textwidth, trim={2.2cm 5.3cm 3.15cm 1.6cm},clip]{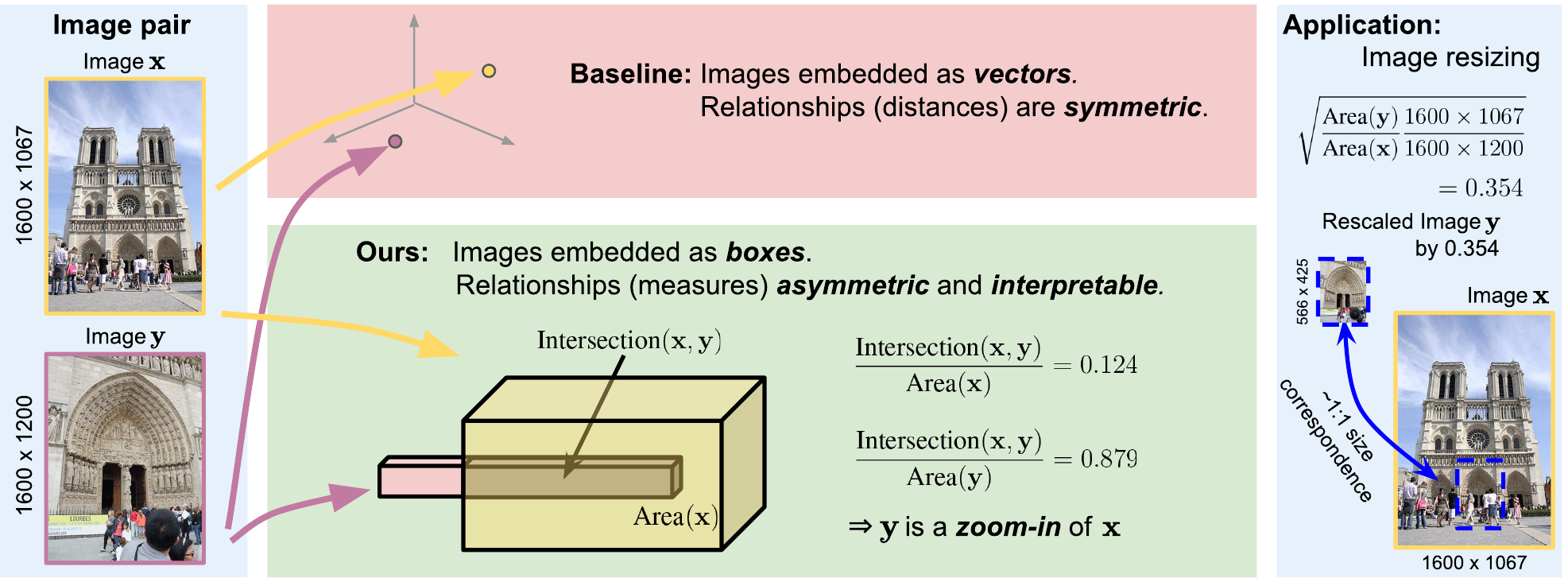}
    \caption{Given two images from the test set, can we reason about their relationship? How much of the scene is visible in both images? Is one image a zoomed-in version of another? We propose a CNN model to predict box embeddings that approximate the visible surface overlap measures between two images. The surface overlap measures give an interpretable relationship between two images and their poses. Here, we extract the relative scale difference between two test images, without expensive geometric analysis. 
    }
    \label{fig:teaser}
\end{figure}

Given two images of the same scene, which one is the close-up? This question is relevant for many tasks, such as image-based rendering %
and navigation~\cite{SGSS-siggraph08}, because the close-up has higher resolution details~\cite{mikulik2014mining} for texturing a generated 3D model, while the other image has the context view of the scene. 
Related tasks include 3D scene reconstruction~\cite{schoenberger2016sfm}, robot navigation~\cite{bonin2008visual} and %
relocalization~\cite{sattler2012image,sattler2011fast,balntas2018relocnet,sattler2016efficient}, which all need to reason in 3D about visually overlapping images. 

For these applications, exhaustive searches are extremely expensive, so efficiency is needed in two related sub-tasks. First, it is attractive to cheaply identify which image(s) from a corpus have substantial overlap with a query image. Most images are irrelevant~\cite{snavely2006photo}, so progress on whole-image descriptors like~\cite{shen2018matchable} narrows that search.
Second, for every two images that are expected to have matchable content, the geometric verification of relative pose involves two-view feature matching and pose estimation~\cite{hartley1997defense}, which can range from being moderately to very expensive~\cite{frahm2010building}, depending on design choices. RANSAC-type routines are more accurate and faster if the two images have matching scales~\cite{zhou2017progressive,Dufournaud:2004,Witkin1983ScaleSpaceF}, and if the detector and descriptor work well with that scene to yield low outlier-ratios.

For scenes where training data is available, we efficiently address both sub-tasks, relevant image-finding and scale estimation, with our new embedding. Our proposed model projects whole images to our custom asymmetric feature space. There, we can compare the non-Euclidean measure between image encoding $\mathbf{x}$ to image encoding $\mathbf{y}$, which is different from comparing $\mathbf{y}$ to $\mathbf{x}$ -- see Figure~\ref{fig:teaser}. %
Ours is distinct from previous methods which proposed a) learning image similarity with metric learning, \eg~\cite{shen2018matchable,arandjelovic2016netvlad,balntas2018relocnet}, and b) estimation of relative scale or image overlap using geometric validation of matched image features, \eg~\cite{mikulik2013browsing,mikulik2014mining}.  
Overall, we
\begin{itemize}
\item advocate that normalized surface overlap (NSO) serves as an interpretable real-world measure of how the same geometry is pictured in two images,
    \item propose a new box embedding for images, that approximates the surface overlap measure while preserving interpretability, and
    \item show that the predicted surface overlap allows us to pre-scale images for same-or-better feature matching and pose estimation in a localization task.
\end{itemize} 

The new representation borrows from box embeddings~\cite{li2018smoothing, vilnis2018probabilistic,subramanian2018new}, that were designed to represent hierarchical relations in words. We are the first to adapt them for computer vision tasks, and hereby propose \emph{image} box embeddings. We qualitatively validate their ability to yield \emph{interpretable} image relationships that don't impede localization, and quantitatively demonstrate that they help with scale estimation.

\section{Related Work}
Ability to quickly determine if two images observe the same scene is useful in multiple computer vision pipelines, especially for SLAM, Structure-from-Motion (SfM), image-based localization, landmark recognition, \etc.

For example, Bag-of-Words (BoW) models~\cite{GalvezTRO12} are used in many SLAM systems, \eg ORB-SLAM2~\cite{orbslam2}, to detect when the camera has revisited a mapped place, \ie a loop closure detection, to re-optimize the map of the scene. 
The BoW model allows to quickly search images of a mapped area and find a match to the current frame for geometric verification.
A similar use case for BoW model can be found in SfM pipelines~\cite{schoenberger2016sfm,frahm2010building,heinly2015reconstructing}, where thousands or even millions of images are downloaded from the Internet and two-view matching is impractical to do between all pairs of images.
Once a 3D model of a scene is built, a user might want to browse the collection of images~\cite{snavely2006photo}. Finding images that are zoom-ins to parts of the query and provide detailed, high-resolution images of interesting parts of the scene was addressed in~\cite{mikulik2013browsing,mikulik2014mining}, where they modified a BoW model to also store local feature geometry information~\cite{perd2009efficient} that can be exploited for Document at a Time (DAAT) scoring~\cite{stewenius2012size} and further geometric verification. However, that approach relies on iterative query expansion to find zoomed-in matches, and re-ranking of retrieved images still requires geometric verification, even when choosing images for query expansion.
Similarly the loss of detail in SfM reconstructions was identified to be a side-effect of retrieval approaches in~\cite{schonberger2015single} and solution based on modification of query expansion and DAAT scoring was proposed, that favors retrieval of images with scale and viewing directions changes. Again, each image query retrieves a subset of images from the image collection in a certain ranking order and geometric verification is performed to establish the pairwise relation of the views. %
Weyand and Leibe~\cite{iconoid1} build a hierarchy of iconic views of a scene by iteratively propagating and verifying homography between pairs of images and generating clusters of views. Later, they extended~\cite{iconoid2} it to find details and scene structure with hierarchical clustering over scale space. Again, the relation between views is estimated with geometric verification.

Sch\"{o}nberger~\etal~\cite{schonberger2015paige} propose to learn to identify image pairs with scene overlap without geometric verification for large-scale SfM reconstruction applications to decrease the total computation time. However, their random forest classifier is trained with binary labels, a pair of images either overlap or not, and no additional output about the relative pose relation is available without further geometric verification.
Shen~\etal~\cite{shen2018matchable} proposed to learn an embedding which can be used for quick assessment of image overlap for SfM reconstructions. They trained a neural network with triplet loss with ground-truth overlap triplets, however their embedding models minimal overlap between two images, which can be used to rank images according to predicted mutual overlap, but does not provide additional information about relative scale and pose of the image pairs.

Neural networks can be used to directly regress a homography between a pair of images~\cite{nguyen2018unsupervised,le2020deep,detone2016deep,erlik2017homography}, however a homography only explains a single planar surface co-visible between views.

Finally, image retrieval is a common technique in localization~\cite{sattler2012image,sattler2017large,balntas2018relocnet}: first, a database of geo-tagged images is searched to find images that are similar to the query, followed by image matching and pose estimation to estimate the pose of the query. Again, the image retrieval reduces the search space for local feature matching, and it is done using a compact representation such as BoW~\cite{philbin2007object}, VLAD~\cite{arandjelovic2013all} or Fisher vectors~\cite{perronnin2010large}. Recently proposed methods use neural networks to compute the image representation and learn the image similarity using geometric information available for the training images, e.g. GPS coordinates for panoramic images~\cite{arandjelovic2016netvlad}, camera poses in world coordinate system~\cite{balntas2018relocnet}, camera pose and 3D model of the environment~\cite{shen2018matchable}.
Networks trained to directly regress the camera pose~\cite{kendall2015posenet,laskar2017camera} were also shown to be similar to image retrieval methods~\cite{sattler2019understanding}. The image embeddings are learned to implicitly encapsulate the geometry of a scene so that similar views of a scene are close in the embedding space. However, these representations are typically learned with metric learning, and so they only allow to rank the database images with respect to the query, and no additional camera pose information is encoded in the representations. We should also mention that localization can also be tackled by directly finding correspondences from 2D features in the query to 3D points in the 3D model without explicit image retrieval, \eg ~\cite{sattler2012improving,brachmann2018learning,sattler2016efficient,sattler2011fast}.

\paragraph{Image Matching}
If a pair of images have a large viewing angle and scale difference, then estimating relative pose using standard correspondence matching becomes challenging. MODS~\cite{mishkin2015mods} addresses the problem by iteratively generating and matching warped views of the input image pair, until a sufficient confidence in the estimated relative pose is reached. Zhou~\etal~\cite{zhou2017progressive} tackles the problem of very large scale difference between the two views by exploiting the consistent scale ratio between matching feature points. Thus, their two-stage algorithm first does exhaustive scale level matching to estimate the relative scale difference between the two views and then does feature matching only in corresponding scale levels. The first stage is done exhaustively, as there is no prior information about the scale difference, which our embeddings provide an estimate for. 

\paragraph{Metric Learning}
Many computer vision applications require learning an embedding, such that the relative distances between inputs can be predicted. For example, face recognition~\cite{schroff2015facenet}, descriptor learning~\cite{he2018local, revaud2019r2d2, tian2019sosnet, NIPS2017-7068, balntas2016learning}, image retrieval~\cite{gordo2016deep}, person re-identification~\cite{hermans2017defense}, \etc. The common approach is metric learning: models are learned to encode the input data (\eg image, patch, \etc) as a vector, such that a desirable distance between input data points is approximated by distances in corresponding vector representations. Typically, the setup uses siamese neural networks, and contrastive loss~\cite{hadsell2006dimensionality}, triplet loss~\cite{schroff2015facenet}, average precision loss~\cite{he2018local} and ranking loss~\cite{Cakir-CVPR} is used to train the network.
The distances in the embedding space can be computed as Euclidean distance between two vectors, an inner product of two vectors, a cosine of the angle between two vectors, \etc. However, learned embeddings can be used to estimate the order of data points, but not other types of relations, due to symmetric distance function used for similarity measure. Other relations, for example, hierarchies require asymmetric similarity measures, which is often encountered in word embeddings~\cite{vilnis2014word, li2018smoothing,vilnis2018probabilistic,subramanian2018new, NIPS2017-7213,vendrov2015order}.

Some methods learn representations that are disentangled, so that they correspond to meaningful variations of the data. For example~\cite{kulkarni2015deep, Worrall_2017_ICCV, sohn2012learning} model geometric transformations of the data in the representation space. However these representations are typically learned for a single class of objects, and it is not straight-forward how to use these embeddings for retrieval.

\section{Method}\label{sec:method}
Our aim is to i) interpret images according to a geometric world-space measure, and ii) to devise an embedding that, once trained for a specific reconstructed $3$D scene, lends itself to easily compute those interpretations for new images. We now explain each, in turn. 
\subsection{World-space Measures}
\begin{figure}[t]
    \centering
    \includegraphics[width=\textwidth, trim={0cm 26.3cm 3.25cm 0},clip]{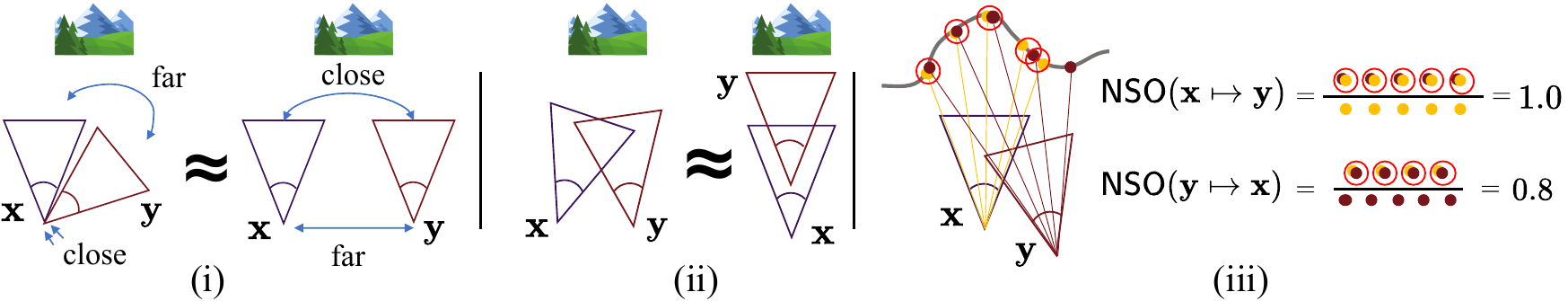}
    \caption{Illustrations of world-space measures and their properties. (i) Left and right camera configurations result in the same value for weighted sum of rotation and translation errors. (ii) Left and right camera configurations result in the same frustum overlap value. (iii) Illustration of our surface overlap. Images $\mathbf{x}$ and $\mathbf{y}$ have resolution of 5 pixels. All 5 pixels of $\mathbf{x}$ are visible in $\mathbf{y}$, meaning the corresponding 3D points of $\mathbf{x}$ are sufficiently close to 3D points backprojected from $\mathbf{y}$. Hence $\mathsf{NSO}(\mathbf{x} \mapsto \mathbf{y})=1.0$. However, only 80\% of pixels of $\mathbf{y}$ are visible by $\mathbf{x}$, thus, $\mathsf{NSO}(\mathbf{y} \mapsto \mathbf{x})=0.8$.}
    \label{fig:world-space-measures}
\end{figure}

We want to have interpretable world-space relationship between a pair of images $\mathbf{x}$ and $\mathbf{y}$, and their corresponding camera poses (orientations and positions). 

A straightforward world-space measure is the distance between camera centers~\cite{arandjelovic2016netvlad}. This world-space measure is useful for localizing omni-directional or panoramic cameras that observe the scene in $360^{\circ}$.
However, most cameras are not omni-directional, so one needs to incorporate the orientation of the cameras. But how can we combine the relative rotation of the cameras and their relative translation into a world-space measure? One could use a weighted sum of the rotation and translation differences~\cite{Buehler:2001,SGSS-siggraph08,snavely2006photo,kendall2015posenet}, but there are many camera configurations where this measure is not satisfactory, \eg~Figure~\ref{fig:world-space-measures}(i).

Another example of a world-space measure is frustum overlap~\cite{balntas2018relocnet}. Indeed, if we extend the viewing frustum of each camera up to a cutoff distance $D$, we can assume that the amount of frustum overlap correlates with the positions and orientation of the cameras. However, the two cameras can be placed multiple ways and have the same frustum overlap -- see Figure~\ref{fig:world-space-measures}(ii). So, the normalized frustum overlap value does not provide interpretable information about the two camera poses.

We propose to use normalized surface overlap for the world-space measure. See Figure~\ref{fig:world-space-measures}(iii) for an illustration. Formally, normalized surface overlap is defined as 
\begin{align}
    \mathsf{NSO}(\mathbf{x} \mapsto \mathbf{y}) &= {\mathsf{overlap}(\mathbf{x} \mapsto \mathbf{y})} / {N_\mathbf{x}}, \textnormal{ and } \label{eq:soxy} \\
    \mathsf{NSO}(\mathbf{y} \mapsto \mathbf{x}) &= {\mathsf{overlap}(\mathbf{y} \mapsto \mathbf{x})} / {N_\mathbf{y}}, \label{eq:soyx}
\end{align}
where $\mathsf{overlap}(\mathbf{x} \mapsto \mathbf{y})$ is the number of pixels in image $\mathbf{x}$ that are visible in image $\mathbf{y}$, and $\mathsf{NSO}(\mathbf{x} \mapsto \mathbf{y})$ is $\mathsf{overlap}(\mathbf{x} \mapsto \mathbf{y})$ normalized by the number of pixels in $\mathbf{x}$ (denoted by $N_\mathbf{x}$), hence it is a number between 0 and 1 and it can be represented as a percentage.
To compute it, we need to know camera poses and the depths of pixels for both image $\mathbf{x}$ and image $\mathbf{y}$. The pixels in both images are backprojected into 3D. The $\mathsf{overlap}(\textbf{x} \mapsto \textbf{y})$ are those 3D points in $\mathbf{x}$ that have a neighbor in the point cloud of image $\mathbf{y}$ within a certain radius.

The normalized surface overlap is not symmetric, $\mathsf{NSO}(\mathbf{x} \mapsto \mathbf{y}) \neq \mathsf{NSO}(\mathbf{y} \mapsto \mathbf{x})$,  because only a few pixels in image $\mathbf{x}$ could be viewed in image $\mathbf{y}$, but all the pixels in $\mathbf{y}$ could be viewed in image $\mathbf{x}$. This asymmetric measure can have an interpretation that image $\mathbf{y}$ is a \emph{close-up view of a part the scene} observed in image $\mathbf{x}$. So, the normalized surface overlap provides an interpretable relation between cameras; please see Figure~\ref{fig:interpretations} for different cases. %

In addition to the visibility of pixels in $\mathbf{x}$ by image $\mathbf{y}$, one could also consider the angle at which the overlapping surfaces are observed. Thus, we weight each point in $\mathsf{overlap}(\mathbf{x} \mapsto \mathbf{y})$ with $\cos(\mathbf{n_i}, \mathbf{n_j})$%
, $\mathbf{n_i}$ denotes the normal of a pixel $\textbf{i}$ and $\mathbf{n_j}$ is the normal of the nearest 3D point of $\textbf{i}$ in image $\textbf{y}$. 
 This will reduce the 
surface overlap
between images observing the same scene from very different angles. Two images are difficult to match if there is substantial perspective distortion due to a difference in viewing angle. Incorporating the angle difference into the world-space measure captures this information.

\subsection{Embeddings}
We aim to learn a representation and embedding of RGB images. The values computed on those representations should approximate the normalized surface overlap measures, but without access to pose and depth data. Such a representation and embedding should provide estimates of the surface overlap measure cheaply and generalize to test images for downstream tasks such as localization.

\paragraph{Vector Embeddings}
A common approach in computer vision to learn embeddings is to use metric learning. Images are encoded as vectors using a CNN, and the network is trained to approximate the relation in world-space measure by distances in vector space. %

However, there is a fundamental limitation of vector representations and metric learning: the distance between vectors corresponding to images $\mathbf{x}$ and $\mathbf{y}$ can only be symmetric. In our case, the normalized surface overlaps are not symmetric. Hence they cannot be represented with vector embeddings. We could compromise and compute a symmetric value, for example the average of normalized surface overlaps computed in both directions,
\begin{equation}
     \mathsf{NSO}^{sym}(\mathbf{x}, \mathbf{y}) = \mathsf{NSO}^{sym}(\mathbf{y}, \mathbf{x}) =
     \frac{1}{2} (\mathsf{NSO}(\mathbf{x} \mapsto \mathbf{y}) + \mathsf{NSO}(\mathbf{y} \mapsto \mathbf{x})).
\end{equation}
Or, only consider the least amount of overlap in one of the directions~\cite{shen2018matchable}, so $\mathsf{NSO}^{min}(\mathbf{x}, \mathbf{y}) =  \mathsf{NSO}^{min}(\mathbf{y}, \mathbf{x}) = \min (\mathsf{NSO}(\mathbf{x} \mapsto \mathbf{y}), \mathsf{NSO}(\mathbf{y} \mapsto \mathbf{x}))$. As shown in Figure~\ref{fig:symmetric-overlap} any such compromise would result in the loss of interpretability.

Different strategies to train vector embeddings are valid. One could consider training a symmetric version using $\mathsf{NSO}^{sym}$ or an asymmetric one using $\mathsf{NSO}$; we hypothesize both are equivalent for large numbers of pairs. We empirically found the best performance with the loss function $\mathcal{L_{\textnormal{vector}}} = ||\mathsf{NSO}^{sym}(\mathbf{x}, \mathbf{y}) - 1 - ||f(\mathbf{x}) - f(\mathbf{y})||_2||_2$, where $f$ is a network that predicts a vector embedding. 

\paragraph{Box Embeddings}
We propose to use box representations to embed images with non-metric world-space measures. Our method is adapted from geometrically-inspired word embeddings for natural language processing~\cite{li2018smoothing, vilnis2018probabilistic,subramanian2018new}. 
The box representation of image $\mathbf{x}$ is a $D$-dimensional orthotope (hyperrectangle) parameterized as a $2D$-dimensional array $\mathbf{b}_\mathbf{x}$. The values of this array are split into $m_{\mathbf{b}_\mathbf{x}}$ and $M_{\mathbf{b}_\mathbf{x}}$, which are the lower and upper bounds of the box in $D$-dimensional space.
Crucially, the box representation allows us to compute the \emph{intersection} of two boxes, $\mathbf{b}_\mathbf{x} \wedge \mathbf{b}_\mathbf{y}$, as
\begin{equation}
\mathbf{b}_\mathbf{x} \wedge \mathbf{b}_\mathbf{y} = \prod_d^D
\sigma \left(\min(M_{\mathbf{b}_\mathbf{x}}^d, M_{\mathbf{b}_\mathbf{y}}^d) - \max(m_{\mathbf{b}_\mathbf{x}}^d, m_{\mathbf{b}_\mathbf{y}}^d)\right),
\end{equation}
\noindent and the \emph{volume} of a box, %
\begin{equation}
 A(\mathbf{b}_\mathbf{x}) = \prod^D_d \sigma(M_{\mathbf{b}_\mathbf{x}}^d- m_{\mathbf{b}_\mathbf{x}}^d),
\end{equation}
where $\sigma(v) = \max(0, v)$. This definition of $\sigma()$ has a problem of zero gradient to $v$ for non-overlapping boxes that should be overlapping. As suggested by Li~\etal~\cite{li2018smoothing}, we train with a smoothed box intersection: $\sigma_\texttt{smooth}(v) = \rho \ln(1 + \exp(v / \rho ))$, which is equivalent to  $\sigma()$ as $\rho$ approaches 0. Hence, we can approximate world-space surface overlap values with normalized box overlaps as
\begin{align}
    \mathsf{NBO}(\mathbf{b}_\mathbf{x} \mapsto \mathbf{b}_\mathbf{y}) = \frac{\mathbf{b}_\mathbf{x} \wedge \mathbf{b}_\mathbf{y}}{A(\mathbf{b}_\mathbf{x})} &\approx \mathsf{NSO}(\mathbf{x} \mapsto \mathbf{y})
     \textnormal{ and }  \\
        \mathsf{NBO}(\mathbf{b}_\mathbf{y} \mapsto \mathbf{b}_\mathbf{x}) = \frac{\mathbf{b}_\mathbf{y} \wedge \mathbf{b}_\mathbf{x}}{A(\mathbf{b}_\mathbf{y})} &\approx \mathsf{NSO}(\mathbf{y} \mapsto \mathbf{x})
    .
\end{align}
For training our embeddings, we can minimize the squared error between the ground truth surface overlap and the predicted box overlap of a random image pair $(\mathbf{x}, \mathbf{y})$ from the training set, so the loss is
\begin{equation}
    \mathcal{L_{\textnormal{box}}} = || \mathsf{NSO}(\mathbf{x} \mapsto \mathbf{y}) - \mathsf{NBO}(\mathbf{b}_\mathbf{x} \mapsto \mathbf{b}_\mathbf{y}) ||^2_2.
\end{equation}
It is important to note that computing the volume of a box is computationally very efficient. Given two boxes, intersection can be computed with just min, max and multiplication operations. Once images are embedded, one could build an efficient search system using R-trees~\cite{guttman1984r} or similar data structures.
\subsection{Implementation Details}
We use a pretrained ResNet50~\cite{he2016deep} as a backbone. The output of the 5-th layer is average pooled across spatial dimensions to produce a 2048-dimensional vector, followed by two densely connected layers with feature dimensions 512 and $2D$ respectively, where $D$ denotes the dimensionality of the boxes. The first $D$ values do not have a non-linearity and represent the position of the center of the box, and the other $D$ values have softplus activation, and represent the size of the box in each dimension. These values are trivially manipulated to represent $m_{\mathbf{b}_\mathbf{x}}$ and $M_{\mathbf{b}_\mathbf{x}}$. We found $D=32$ to be optimal (ablation in supp. material). Input images are anisotropically rescaled to $256 \times 456$ resolution.  We fix $\rho=5$ when computing smoothed box intersection and volume. We train with batch size $b=32$ for $20-60k$ steps, depending on the scene. 
 The ground-truth overlap both for training and evaluation is computed over the original resolution of images, however the 3D point cloud is randomly subsampled to 5000 points for efficiency. Surface normals are estimated by fitting a plane to a $3\times3$ neighborhood of depths. We used $0.1$ as the distance threshold for two 3D points to overlap.

\section{Experiments}
We conduct four main experiments.
First, we validate how our box embeddings of images capture the surface overlap measure. Do these embeddings learn the asymmetric surface overlap and preserve interpretability? Do predicted relations of images match camera pose relations?
 
Second, we demonstrate that the proposed world measure is superior to alternatives like frustum overlap for the task of image localization. 
We evaluate localization quality on a small-scale indoor dataset and a medium-scale outdoor dataset. We also show that images retrieved with our embeddings can be ranked with metric distances, and are ``backwards-compatible'' with existing localization pipelines, despite also being interpretable.

Third, we demonstrate how the interpretability of our embeddings can be exploited for the task of localization, specifically for images that have large scale differences to the retrieved gallery images. Finally, we show how our embeddings give scale estimates that help with feature extraction.

\paragraph{Datasets}%
Since our proposed world-space measure requires depth information during training, we conduct our experiments on MegaDepth~\cite{MegaDepthLi18} and 7Scenes~\cite{shotton2013scene} datasets.

MegaDepth is a large scale dataset including 196 scenes, each of which consists of images with camera poses and camera intrinsics. However, only a subset of images have a corresponding semi-dense depth map. The dataset was originally proposed for the task of SfM reconstruction and to learn single-image depth estimation methods. The scenes have varying numbers of images, and large scenes have camera pose variations with significant zoom-in and zoom-out relationships. Most images do not have corresponding semi-dense depth maps, and may have only two-pixel ordinal labels or no labels at all. The low number of images with suitable depth also necessitates the generation of our own train and test splits, replacing the provided sets that are sampled across images with and without depths. As depth is needed for the computation of the ground-truth surface overlap, we only consider four scenes %
that provide enough data for training, validation and testing. So, we leave 100 images for validation and 100 images for testing, and the remaining images with suitable depth maps are used for training: 1336, 2130, 2155 and 1471 images for Big Ben, Notre-Dame, Venice, and Florence scenes, respectively.

7Scenes is an established small-scale indoor localization benchmark consisting of 3 to 10 sequences of a thousand images (per scene) with associated depth maps Kinect-acquired ground-truth poses. To evaluate our localization pipeline we follow the training and test splits from \cite{shotton2013scene} and compare to methods such as RelocNet~\cite{balntas2018relocnet}, which uses frustum overlap as a world-space measure.

\subsection{Learning Surface Overlap}
Figure~\ref{fig:query-notre-dame} %
shows qualitative results of ground-truth surface overlaps and the predicted box overlaps between a random test image as query and training images as gallery. See supplementary materials for more examples.
\renewcommand{\thefootnote}{\arabic{footnote}}

In Table~\ref{tab:errs} we compare vector embeddings against box embeddings to experimentally validate that asymmetric surface overlap cannot be learned with a symmetric vector embedding. We evaluate the predictions against ground truth surface overlap on $1,000$ random pairs of test images. For each pair of images, we can compute ground truth $\mathsf{NSO}(\mathbf{x} \mapsto \mathbf{y})$ and $\mathsf{NSO}(\mathbf{y} \mapsto \mathbf{x})$. We report results on three metrics: $L_1$-Norm\footnote{
$L_1$-Norm: $\frac{1}{N}\sum |\mathsf{NSO}(\mathbf{x} \mapsto \mathbf{y}) - \mathsf{NBO}(\mathbf{b}_\mathbf{x} \mapsto \mathbf{b}_\mathbf{y})| + |\mathsf{NSO}(\mathbf{y} \mapsto \mathbf{x}) 
    - \mathsf{NBO}(\mathbf{b}_\mathbf{y} \mapsto \mathbf{b}_\mathbf{x})|.$}, the root mean square error (RMSE)\footnote{RMSE:
$\sqrt{\frac{1}{N}\sum [(\mathsf{NSO}(\mathbf{x} \mapsto \mathbf{y}) - \mathsf{NBO}(\mathbf{b}_\mathbf{x} \mapsto \mathbf{b}_\mathbf{y}))^2 + (\mathsf{NSO}(\mathbf{y} \mapsto \mathbf{x}) - \mathsf{NBO}(\mathbf{b}_\mathbf{y} \mapsto \mathbf{b}_\mathbf{x}))^2]}
.
$} and the prediction accuracy. The prediction accuracy is defined as the percentage of individual overlaps that is predicted with an absolute error of less than 10\%. Note, that the ground-truth depth information for the images may be incomplete, so there is inherent noise in the training signal and ground-truth measurements which makes smaller percentage thresholds not meaningful. 

These results confirm that box embeddings are better than vectors at capturing the asymmetric world-space overlap relationships between images. 

\subsection{Interpreting Predicted Relations}
\label{sec:relationship}
For these experiments, we learn box embeddings for images in different scenes of the Megadepth dataset~\cite{MegaDepthLi18}. %
A trained network can be used to predict box representations for any image. We re-use our training data (which incidentally also have ground-truth depths) to form the gallery. Each query image $\mathbf{q}$ comes from the test-split. We compute its box representation $\mathbf{b}_\mathbf{q}$ and compute surface overlap approximations using box representations as specified in Eq.~(\ref{eq:soxy}) and (\ref{eq:soyx}).

\begin{table}[!t]
\sbox\tmpbox{%
    \centering
    \includegraphics[trim={5.2cm 8.7cm 7.43cm 4.3cm},clip,width=0.35\textwidth]{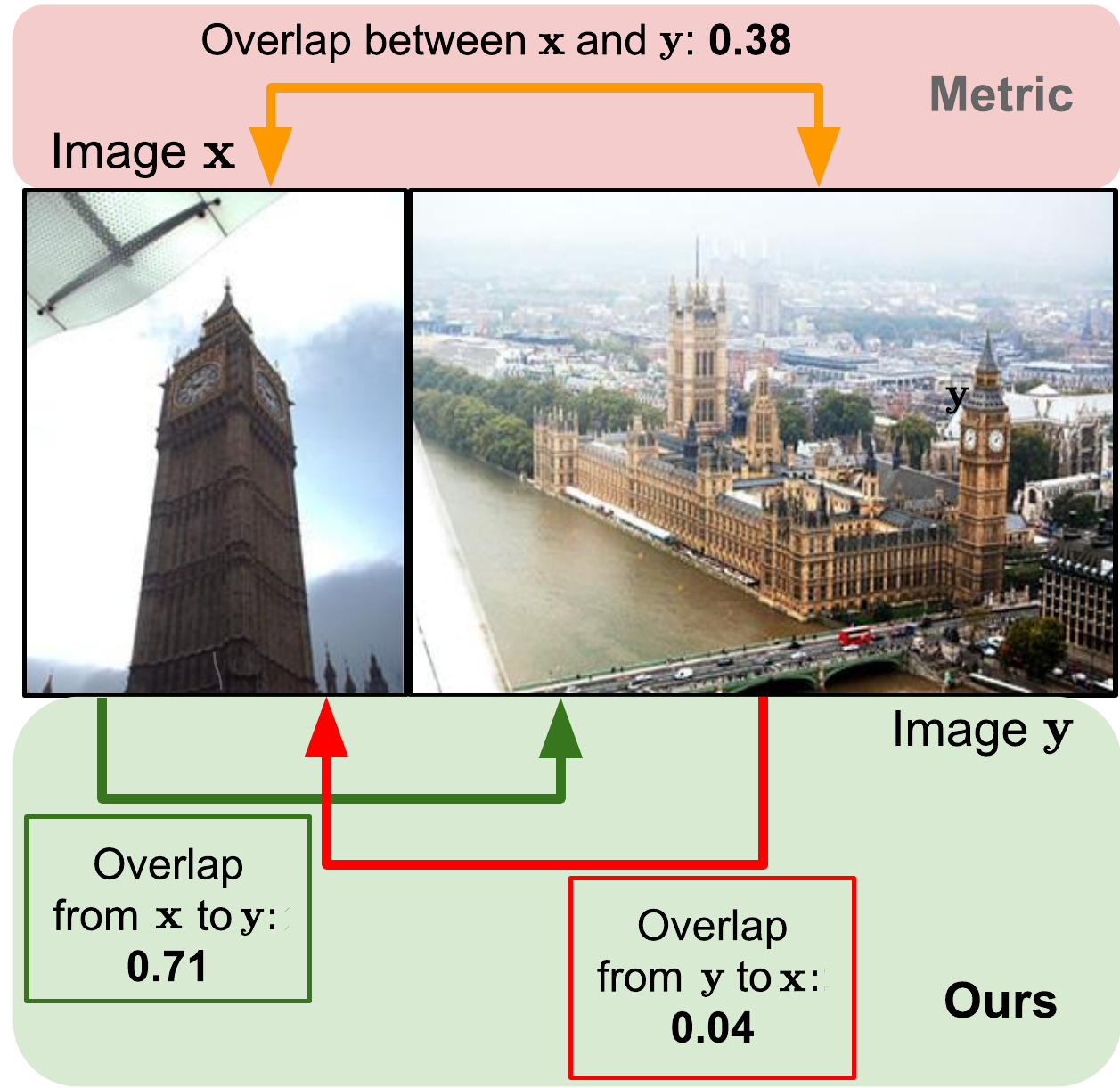}
  }%
  \renewcommand*{\arraystretch}{0}
  \begin{tabular*}{\linewidth}{@{\extracolsep\fill}p{\wd\tmpbox}p{75mm}@{}}
    \usebox\tmpbox &
        \begin{tabular}[b]{c|c|c|c|m{20mm}}
    \centered{\textbf{x}} & \centered{\textbf{y}} &
    \shortstack{\scriptsize $\mathsf{NSO}$\\\scriptsize$(\mathbf{x} \mapsto \mathbf{y})$ }& 
     \shortstack{\scriptsize $\mathsf{NSO}$\\\scriptsize$(\mathbf{y} \mapsto \mathbf{x})$ }&
     Interpretation
    \\
    \hline
    \centered{\includegraphics[width=0.06\textwidth]{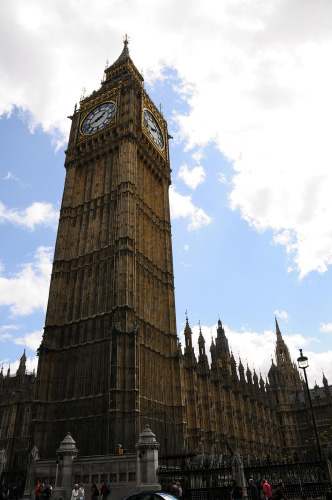}} & \centered{\includegraphics[width=0.11\textwidth]{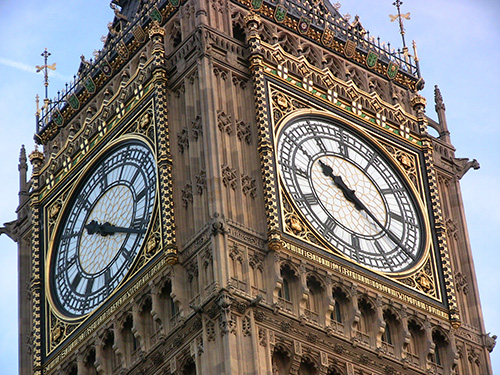}} & \centered{\shortstack{\color{pred}15.2\%\\\color{gt}\tiny (15.3\%)}} & \centered{\shortstack{\color{pred}83.1\%\\\tiny \color{gt}(83.5\%)}} &
     \shortstack{$\mathbf{y}$ is zoom-in\\on $\mathbf{x}$}
     \\
     \centered{\includegraphics[width=0.11\textwidth]{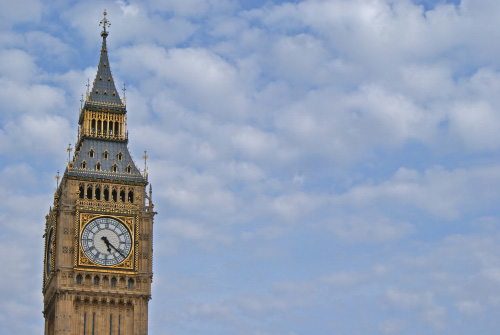}} & \centered{\includegraphics[width=0.06\textwidth]{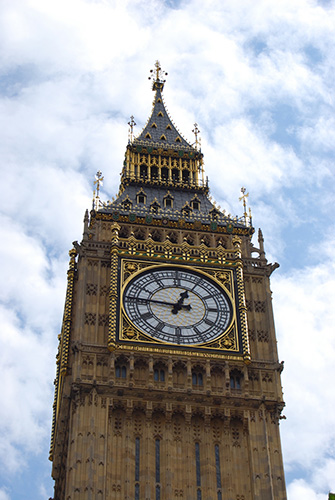}} &  \centered{\shortstack{\color{pred}80.8\%\\\color{gt}\tiny (87.7\%)}} & \centered{\shortstack{\color{pred}88.7\%\\\color{gt}\tiny (89.2\%)} }
     & 
     \shortstack{$\mathbf{x}$ and $\mathbf{y}$ are\\clone-like}
     \\

 \centered{\includegraphics[width=0.11\textwidth]{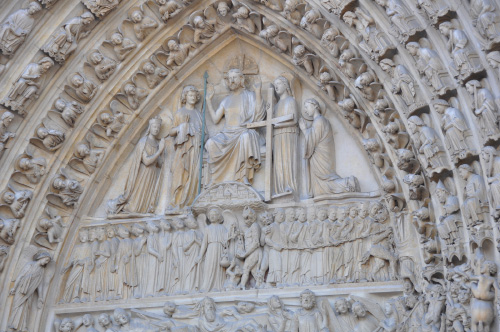}} & 
 \centered{\includegraphics[width=0.06\textwidth]{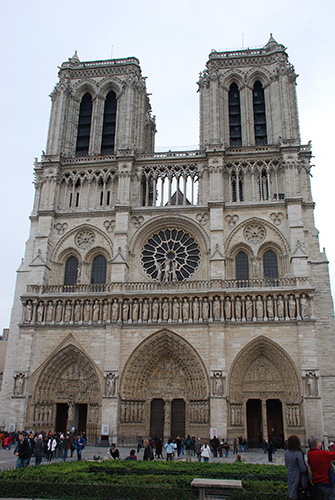}} & \centered{\shortstack{\color{pred}85.3\%\\\color{gt}\tiny (86.9\%)}} & \centered{\shortstack{\color{pred}5.3\%\\\color{gt}\tiny (4.8\%)}}
 & \shortstack{$\mathbf{y}$ is zoom-out\\of $\mathbf{x}$}
 \\ 
 
\centered{\includegraphics[width=0.06\textwidth]{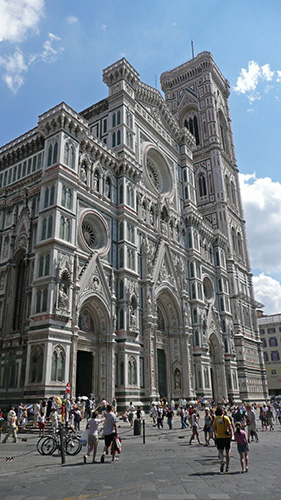}}&
\centered{\includegraphics[width=0.11\textwidth]{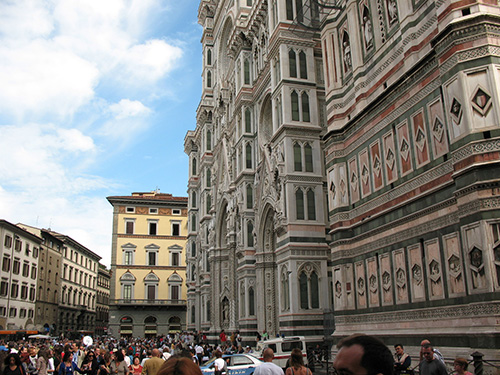}} & \centered{\shortstack{\color{pred}23.7\%\\\color{gt}\tiny (20.1\%)}}  &  \centered{\shortstack{\color{pred}22.3\%\\\color{gt}\tiny (28.5\%)}} &
\shortstack{$\mathbf{y}$ is\\ oblique-out or\\ crop-out of $\mathbf{x}$} \\
    \end{tabular}%
        \\
            \captionof{figure}{Top: Metric learning can only represent a single distance between two images of Big Ben. Bottom: While 4\% of image $\mathbf{y}$ is visible in image $\mathbf{x}$, 71\% of image $\mathbf{x}$ is visible vice versa. The average of overlaps is $38$\%.}
    \label{fig:symmetric-overlap}
    &
                \captionof{figure}{Interpretation of image relationship based on the normalized surface overlap ($\mathsf{NSO}$) between two images (\color{pred} predicted  \color{black} from just RGB with our approach and \color{gt} ground-truth\color{black}~using privileged $3$D data). Four different relationships between image pairs can be observed. When $\mathsf{NSO}(\mathbf{x} \mapsto \mathbf{y})$ is low but $\mathsf{NSO}(\mathbf{y} \mapsto \mathbf{x})$ is high, this indicates that most pixels of $\mathbf{y}$ are visible in $\mathbf{x}$. Therefore $\mathbf{y}$ must be a close-up of $\mathbf{x}$. %
                }
        \label{fig:interpretations}
    \label{fig:image}
  \end{tabular*}
\end{table}

First, we analyze predicted relations qualitatively.
We introduce two terms to discuss the relations between query image $\mathbf{q}$ and retrieved image $\mathbf{r}$: \emph{enclosure} and \emph{concentration}. %
Enclosure refers to the predicted surface overlap from query to retrieved image, \ie $\mathsf{NBO}(\mathbf{b}_\mathbf{q} \mapsto \mathbf{b}_\mathbf{r})$. Concentration refers to the predicted surface overlap from retrieved image to query, \ie $\mathsf{NBO}(\mathbf{b}_\mathbf{r} \mapsto \mathbf{b}_\mathbf{q})$. Thus, if a retrieved image has a large value for enclosure, then it observes a large number of the query's pixels. See Figure~\ref{fig:interpretations} for other interpretations. %

\begin{figure}[!t] %
\begin{minipage}{0.5\textwidth}
\setlength\tabcolsep{0.2pt}
\begin{tabular}{|l|cc|cc@{}|}  %
\hline
&\multicolumn{2}{c}{ High Concentration} & \multicolumn{2}{|c|}{ Low Concentration}\\
\hline
\multirow{6}{*}{\rotatebox[origin=c]{90}{Low Enclosure}}
& \includegraphics[width=0.22\textwidth]{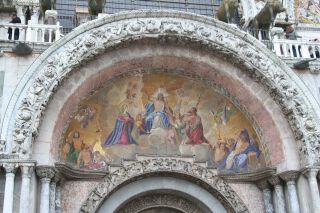}
& \includegraphics[width=0.22\textwidth]{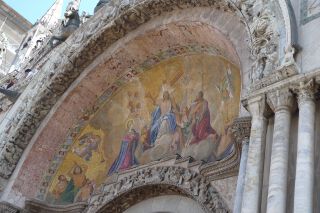}
& \includegraphics[width=0.22\textwidth]{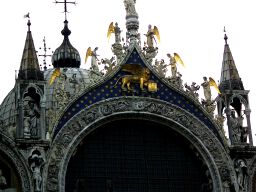}
& \includegraphics[width=0.12\textwidth]{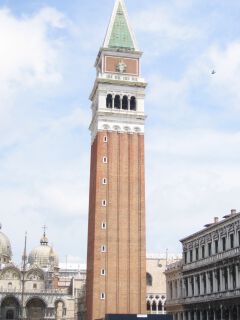}
\\   & \shortstack{\pred{75}{17}\\\gt{87}{33}} & \shortstack{\pred{62}{9}\\ \gt{64}{19}} & \shortstack{\pred{47}{6}\\\gt{14}{2}} &  \shortstack{\pred{8}{14}\\\gt{6}{22}} \\ 

& \includegraphics[width=0.22\textwidth]{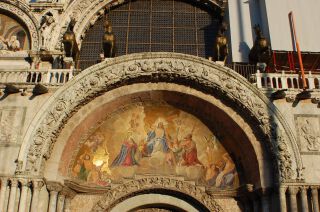}
& \includegraphics[width=0.22\textwidth]{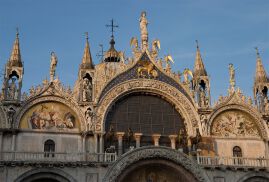}

& \includegraphics[width=0.22\textwidth]{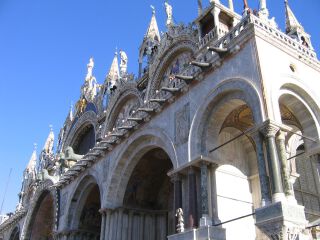}
& \includegraphics[width=0.22\textwidth]{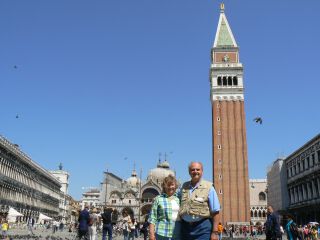}
\\   & \shortstack{\pred{70}{26}\\\gt{73}{32}} & \shortstack{\pred{51}{29}\\\gt{39}{31}} & \shortstack{\pred{32}{25}\\\gt{37}{19}} & \shortstack{\pred{6}{28} \\\gt{2}{21} }\\

& \includegraphics[width=0.22\textwidth]{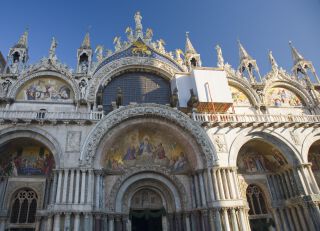}
& \includegraphics[width=0.22\textwidth]{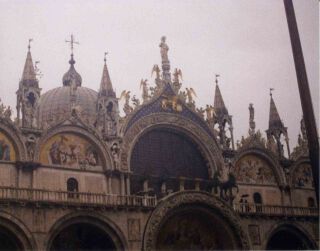}

& \includegraphics[width=0.22\textwidth]{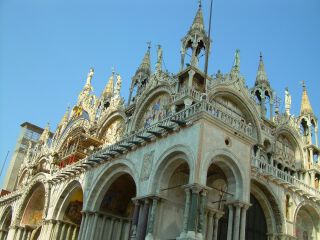}
& \includegraphics[width=0.22\textwidth]{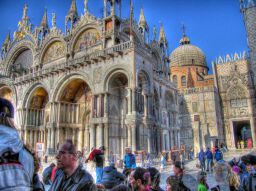}
\\   & \shortstack{\pred{69}{47}\\\gt{66}{63}} & \shortstack{\pred{55}{44}\\\gt{41}{47}} &  \shortstack{\pred{18}{48}\\\gt{21}{33}} & \shortstack{\pred{14}{28}\\\gt{29}{21}}%
\\ 
\hline
\multirow{6}{*}{\rotatebox[origin=c]{90}{High Enclosure}} 
& \includegraphics[width=0.22\textwidth]{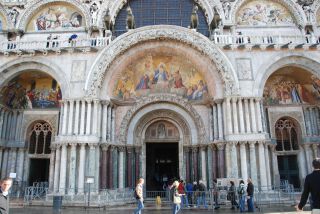} 
& \includegraphics[width=0.22\textwidth]{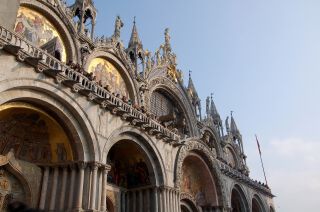}
& \includegraphics[width=0.22\textwidth]{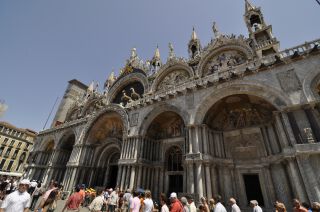}
& \includegraphics[width=0.22\textwidth]{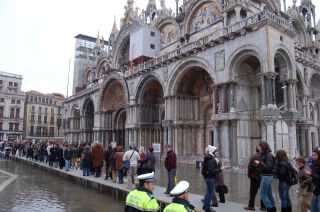}

\\  & \shortstack{\pred{71}{52}\\\gt{76}{69}} & \shortstack{\pred{51}{63}\\\gt{27}{58}} & \shortstack{\pred{45}{63}\\\gt{56}{57}} & \shortstack{\pred{30}{67}\\\gt{31}{45}} 
\\
&\includegraphics[width=0.22\textwidth]{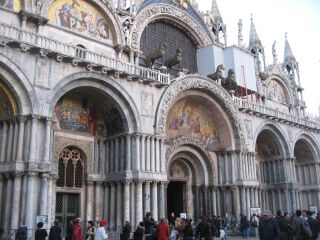}
& \includegraphics[width=0.22\textwidth]{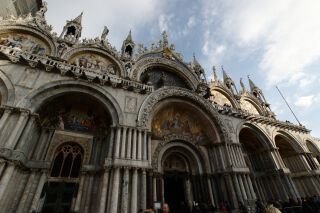}
& \includegraphics[width=0.22\textwidth]{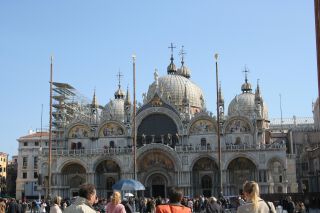}
& \includegraphics[width=0.22\textwidth]{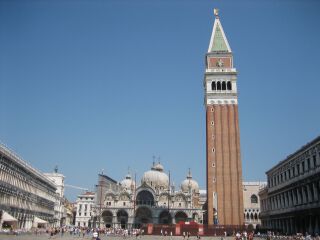}
\\ &  \shortstack{\pred{70}{76}\\\gt{58}{84}} & \shortstack{\pred{60}{77}\\\gt{63}{78}} &  \shortstack{\pred{25}{84}\\\gt{34}{80}} &\shortstack{\pred{6}{75}\\\gt{9}{72}} 
 \\
& \includegraphics[width=0.22\textwidth,cfbox=green 2pt 2pt]{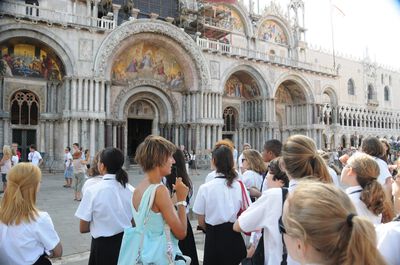}
& \includegraphics[width=0.22\textwidth]{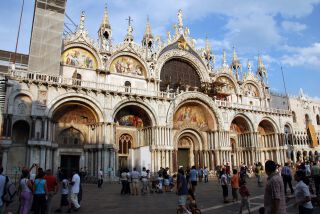}
& \includegraphics[width=0.22\textwidth]{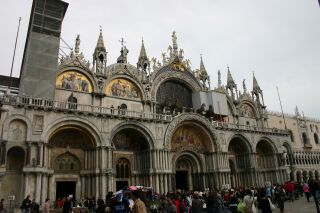}
& \includegraphics[width=0.22\textwidth]{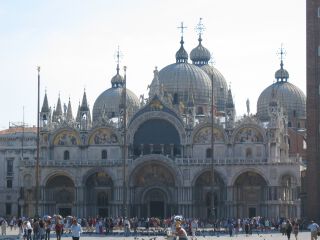}

\\ &\shortstack{query\\image}  & \shortstack{\pred{50}{93}\\\gt{42}{87}} & \shortstack{\pred{45}{89}\\\gt{39}{87}} & \shortstack{\pred{24}{87}\\\gt{32}{81}}   \\  
  \hline
\end{tabular}

\end{minipage}
\begin{minipage}{0.475\textwidth}
\includegraphics[width=0.94\textwidth]{figures/pdf/cluster_test_sfm_sample3.pdf}

\end{minipage}
\caption{Left: Results of predicted and ground-truth enclosure and concentration (defined in Sec~\ref{sec:relationship}) relative to the query image indicated by a green frame.  The numbers below each image indicate the \color{pred}predicted \color{black} and \color{gt}ground-truth \color{black} concentration/enclosure. Right: Results of predicting enclosure and concentration between a query image from the test set and test images (including test images without depth maps). Images are plotted at the coordinates of predicted (enclosure, concentration). Note in both plots how results in the upper right quadrant are oblique views, and bottom-right show zoomed-out views. The upper left quadrants show zoomed-in views, depending on the range of (low) enclosure selected. }
\label{fig:query-notre-dame}
\end{figure}

To demonstrate the interpretability of the predicted relations between the images, we retrieve and show gallery images in different ranges of enclosure and concentration. 
Figure~\ref{fig:query-notre-dame} shows results on two different scenes. On the left we see qualitative and quantitative results for the query image from the test data, and the images retrieved from the training set. 
As can be seen, the images in different quadrants of enclosure and concentration ranges are interpretable with different amounts of zoom-in, or zoom-out, or looking clone-like, or exhibiting crop-out/oblique-out.
On the right, we retrieve images from the larger test set (without depth maps) plotted according to the estimated enclosure and concentration. This qualitative result demonstrates how box embeddings generalize to the test set. 
Please see supplementary materials for more examples. 

\begin{table}[!t]
\begin{center}
\caption{Evaluation of box \vs vector embeddings, trained on normalized surface overlap. We measure the discrepancy between the predicted and ground-truth overlaps on the test set. Across all measures and scenes, the non-metric embeddings with box representations outperform metric learning with vector representations}
\setlength\tabcolsep{4pt}
\begin{tabular}[H]{|l|c|c|c|c|c|c|}
\hline
& \multicolumn{3}{|c}{$\mathsf{NSO}(\mathbf{x} \mapsto \mathbf{y})$ with boxes} 
& \multicolumn{3}{|c|}{$\mathsf{NSO}^{sym}(\mathbf{x},\mathbf{y})$  with vectors } \\ 
\hline
&$L_1$-Norm & RMSE & Acc.$<0.1$ & L1-Norm & RMSE & Acc.$<0.1$ \\
\hline
Notre-Dame & 0.070 & 0.092 & 93.3\%  & 0.244 & 0.249 & 60.9\% \\
Big Ben  & 0.126 &0.138 & 82.6\% & 0.429 & 0.350 & 24.8\% \\
Venice  & 0.066&0.112 & 89.9\% & 0.164 & 0.193 & 75.1\% \\ 
Florence &0.063 & 0.094& 89.7\% & 0.145 & 0.162 & 76.5\% \\
\hline
\end{tabular}
\end{center}
\label{tab:errs}
\end{table}
\subsection{Querying Box Embeddings for Localization}
We now compare the differences between surface overlap and frustum overlap measures for retrieving images in a localization task. 
The task is to find the camera pose of a query image $\mathbf{q}$ with respect to the images in the training set, where the latter have known camera poses and semi-dense depths. The image embeddings are used to retrieve the top $k$-th image ($k=1$) from the training data that are closest according to each embedding measure. %
After retrieval, $2$D$-3$D correspondences are found between the query image and $k$-th retrieved image's 3D point cloud. %
We use SIFT~\cite{lowe2004distinctive} features, and correspondences are filtered with Lowe's ratio test and matched using OpenCV's FLANN-based matcher. The pose is solved with RANSAC and PnP, and measured against the ground truth pose. We report median translation and rotation errors, with all test images as queries.

\begin{table}[t]
\centering
\caption{Results on 7Scenes dataset~\cite{shotton2013scene}. ``Repr.'' indicates the embedding representation as box (B) or vector (V). Q specifies if the world space measure is symmetric (S) or asymmetric (A). Reported numbers show translation and rotation errors in meters and degrees respectively. The results indicate that symmetric surface overlap is superior to frustum overlap when represented with vectors. Asymmetric surface overlap box embeddings are similar to symmetric surface overlap vector embeddings, except for the Stairs scene. The last two rows show the generalization ability of our embeddings: the two embeddings were trained on Kitchen and used for retrieval on other scenes}
\resizebox{\columnwidth}{!}{
\begin{tabular}{|l|c|c|c|c|c|c|c|c|c|}
\hline
Method & {\footnotesize Repr.} & {Q} & Chess & Fire & Heads & Office & Pumpkin & Kitchen & Stairs  \\
\hline
\hline
DenseVLAD \cite{torii201524} &V&S&0.03, 1.40$^{\circ}$  &0.04, 1.62$^{\circ}$  & 0.03, 1.21$^{\circ}$  & 0.05, 1.37$^{\circ}$ & 0.07, 2.02$^{\circ}$  &0.05, 1.63$^{\circ}$ &0.16, 3.85$^{\circ}$ \\
NetVLAD \cite{arandjelovic2016netvlad} &V&S&0.04, 1.29$^{\circ}$ &0.04, 1.85$^{\circ}$ &0.03, 1.14$^{\circ}$ &0.05, 1.45$^{\circ}$ &0.08, 2.16$^{\circ}$ &0.05, 1.77$^{\circ}$  &0.16, 4.00$^{\circ}$ \\
\hline
PoseNet \cite{kendall2015posenet} &&&0.32, 8.12$^{\circ}$ &0.47, 14.4$^{\circ}$ &0.29, 12.0$^{\circ}$ &0.48, 7.68$^{\circ}$ &0.47, 8.42$^{\circ}$ &0.59, 8.64$^{\circ}$ &0.47, 13.6$^{\circ}$ \\
RelocNet \cite{balntas2018relocnet}&&&0.12, 4.14$^{\circ}$ &0.26, 10.4$^{\circ}$ &0.14, 10.5$^{\circ}$ &0.18, 5.32$^{\circ}$ &0.26, 4.17$^{\circ}$ &0.23, 5.08$^{\circ}$ &0.28, 7.53$^{\circ}$ \\
Active Search \cite{sattler2016efficient}&&&0.04, 1.96$^{\circ}$ &0.03, 1.53$^{\circ}$ &0.02, 1.45$^{\circ}$ &0.09, 3.61$^{\circ}$ &0.08, 3.10$^{\circ}$ &0.07, 3.37$^{\circ}$ &0.03, 2.22$^{\circ}$ \\
DSAC++ \cite{brachmann2018learning}&&&\textbf{0.02, 0.50}$^{\circ}$ &\textbf{0.02, 0.90}$^{\circ}$ &\textbf{0.01, 0.80}$^{\circ}$ &\textbf{0.03, 0.70}$^{\circ}$  & \textbf{0.04, 1.10}$^{\circ}$  &\textbf{0.04, 1.10}$^{\circ}$ &\textbf{0.09, 2.60}$^{\circ}$ \\
\hline
\hline
\textbf{Ours $\mathsf{NSO}()$} &B & A & 0.05, 1.47$^{\circ}$ & \textbf{0.05, 1.91$^{\circ}$} & 0.05, 2.54$^{\circ}$  & \textbf{0.06}, 1.60$^{\circ}$  & 0.10, 2.46$^{\circ}$  &0.07, 1.73$^{\circ}$  & 0.50, 9.18$^{\circ}$ \\
   \textbf{Ours $\mathsf{NSO}^{sym}()$} & V & S &  \textbf{0.04, 1.19}$^{\circ}$  &\textbf{0.05}, 2.05$^{\circ}$ & 0.05, 2.84$^{\circ}$  & \textbf{0.06, 1.46}$^{\circ}$  &\textbf{0.10, 2.28}$^{\circ}$   &\textbf{0.06, 1.61}$^{\circ}$  & \textbf{0.22, 5.28}$^{\circ}$ \\

 \textbf{Ours Frustum} & V & S & 0.05, 1.25$^{\circ}$   & \textbf{0.05}, 2.02$^{\circ}$  & \textbf{0.04, 1.86}$^{\circ}$  & 0.07, 1.73$^{\circ}$  &\textbf{0.10}, 2.40$^{\circ}$  & 0.07, 1.71$^{\circ}$  & 1.82, 12.0$^{\circ}$ \\
 \hline\hline
    \textbf{Box (Kitchen)}& B & A  & 0.06, 2.19$^{\circ}$   & 0.08, 2.94$^{\circ}$  & 0.08, 5.42$^{\circ}$  & 0.17, 4.87$^{\circ}$  & 0.13, 3.21$^{\circ}$ & * & 1.83, 50.1$^{\circ}$ \\
  \textbf{Vector (Kitchen)}& V & S & \textbf{0.04}, 1.54$^{\circ}$   & 0.06, 2.26$^{\circ}$  &0.04, 1.90$^{\circ}$   & \textbf{0.06}, 1.88$^{\circ}$  &\textbf{0.10}, 2.37$^{\circ}$  & * & 0.55, 9.22$^{\circ}$ \\
\hline
\end{tabular}
}
\label{tab:seven-scenes}
\end{table}

We compare three embeddings for this task, each trained with a different world-space measure: i) vector embeddings trained with frustum overlap, ii) vector embeddings trained with $\mathsf{NSO}^{sym}()$, and box embeddings trained with $\mathsf{NSO}()$.
Ranking for (i) and (ii) is easy, and for (iii) we rank retrieved images according to $0.5(\mathsf{NBO}(\mathbf{b}_\mathbf{x} \mapsto \mathbf{b}_\mathbf{y}) + \mathsf{NBO}(\mathbf{b}_\mathbf{y}  \mapsto  \mathbf{b}_\mathbf{x}))$. %
This query function is used to show backwards compatibility with traditional metric embeddings.

\paragraph{7Scenes}
Table~\ref{tab:seven-scenes} shows the results %
of using (i) frustum overlap and both (ii) symmetric and  (iii) asymmetric surface overlap. We also report the results of  state-of-the-art work~\cite{kendall2015posenet,balntas2018relocnet,sattler2016efficient, brachmann2018learning} and two SOTA baselines, DenseVLAD and NetVLAD, that we generated by swapping our retrieval system with the ones of \cite{torii201524} and \cite{arandjelovic2016netvlad} respectively and leaving the rest of the pose estimation pipeline intact.  %
Both surface overlap-based results are generally better than frustum overlap. Note, absolute differences between most recent methods on this dataset are relatively minor. For example, due to the use of PnP, we get better results using our frustum overlap than RelocNet. Ours-box is similar in performance to Ours-vector, except for the Stairs dataset, where we perform poorly because stairs have many repeated structures. This is a positive result, showing on-par performance while giving the benefit of an interpretable embedding. Although the localization task includes having a $3$D reconstruction of the gallery/training images by its nature, we also compare two embeddings that were trained on Kitchen and tested generalization for retrieval on other scenes. Please see the supplementary material for further experimental results.

\begin{figure}[b]
    \centering
    \begin{minipage}{0.47\textwidth}
        \centering
        \includegraphics[width=\linewidth]{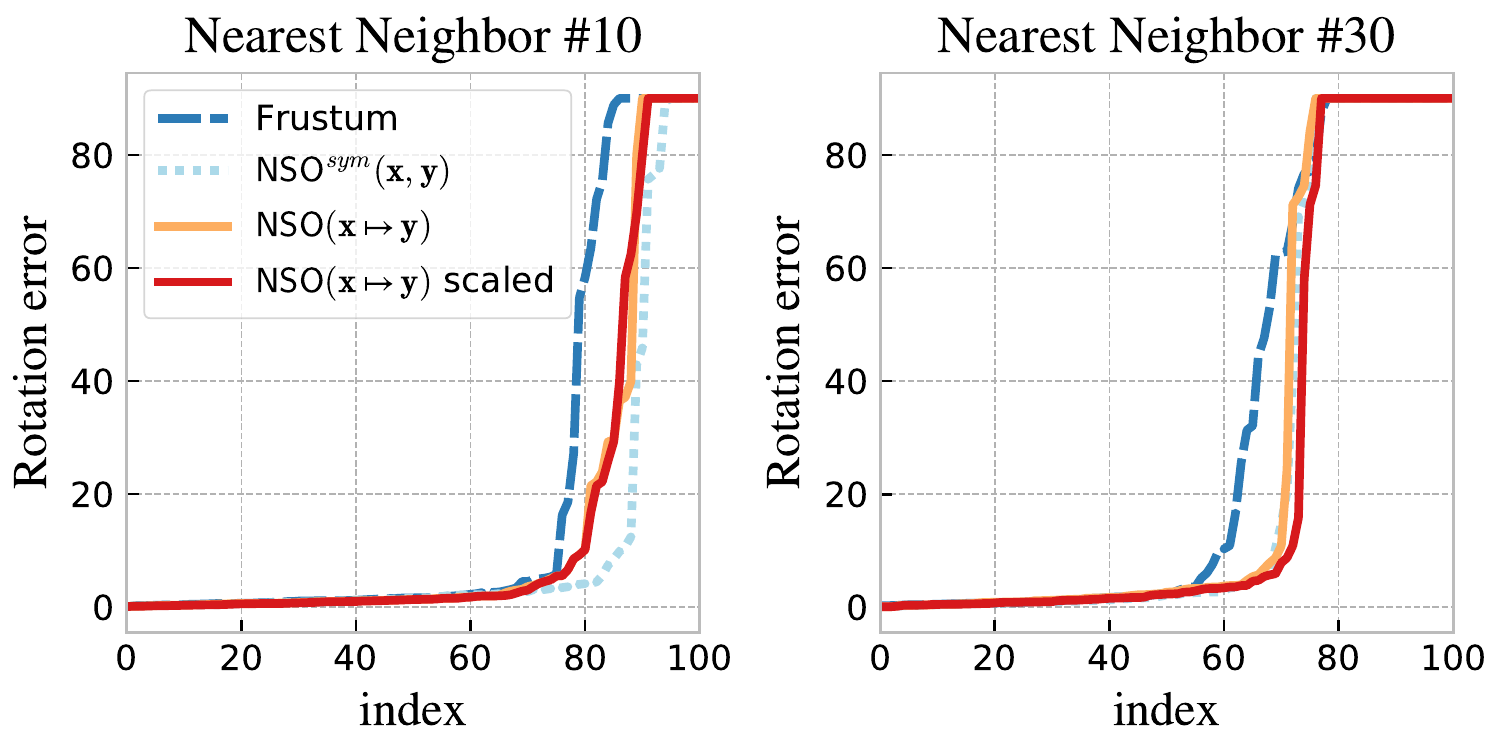}\\
        \footnotesize Notre-Dame
    \end{minipage}
    \begin{minipage}{0.47\textwidth}
        \centering
        \includegraphics[width=\linewidth]{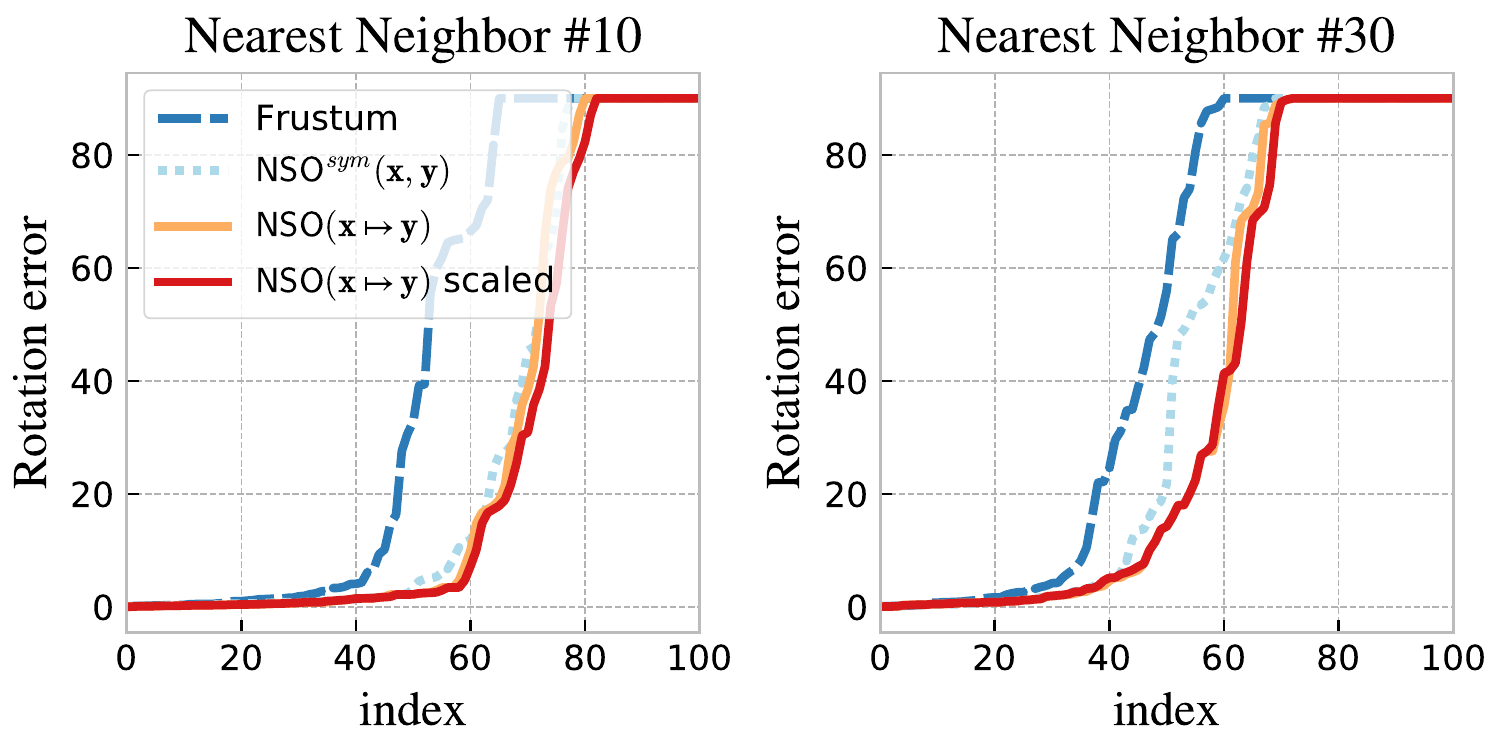}\\
        \footnotesize Big Ben
    \end{minipage}
    \caption{%
    Each plot shows (sorted) rotation error (capped at 90$^\circ$) when each test image is matched against 10-th and 30-th closest retrieved image for pose estimation. As we can see, box embeddings with surface overlap measure tend to outperform alternatives, especially when rescaling images according to estimated relative scale.}
    \label{fig:boxplot}
\end{figure}

\subsection{Predicting Relative Scale Difference for Image Matching}
Given a pair of images $\mathbf{x}$ and $\mathbf{y}$, we can compare the predicted $\mathsf{NBO}(\mathbf{b}_\mathbf{x} \mapsto \mathbf{b}_\mathbf{y})$ and $\mathsf{NBO}(\mathbf{b}_\mathbf{y} \mapsto \mathbf{b}_\mathbf{x})$ to estimate the relative scale of the images, so 
\begin{equation}
\frac{\mathsf{NBO}(\mathbf{b}_\mathbf{x} \mapsto \mathbf{b}_\mathbf{y})}
{\mathsf{NBO}(\mathbf{b}_\mathbf{y} \mapsto \mathbf{b}_\mathbf{x})} 
\approx
\frac{\mathsf{NSO}(\mathbf{x} \mapsto \mathbf{y})}{\mathsf{NSO}(\mathbf{y} \mapsto \mathbf{x})} =
\frac{
     {\mathsf{overlap}(\mathbf{x} \mapsto \mathbf{y})} / {N_\mathbf{x}}
     }{
     {\mathsf{overlap}(\mathbf{y} \mapsto \mathbf{x})} / {N_\mathbf{y}}
     } \approx 
\frac{1}{s^2}
\frac{N_\mathbf{y}}{ N_\mathbf{x}}.
\end{equation}
This estimates the scale factor $s$ to be applied to image $\mathbf{x}$, so that overlaps $\mathsf{overlap}(\mathbf{x} \mapsto \mathbf{y})$ and $\mathsf{overlap}(\mathbf{y} \mapsto \mathbf{x})$ occupy approximately the same number of pixels in each of the two images.

\begin{figure}
\centering
\setlength\tabcolsep{1pt}
\begin{tabular}{c|c|c|c} %
\includegraphics[width=0.21\textwidth]{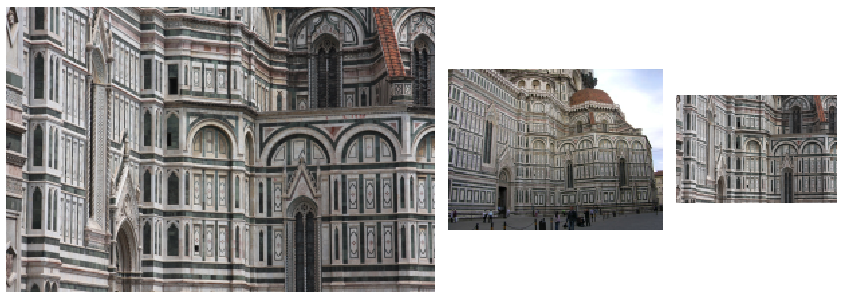} &
\includegraphics[width=0.21\textwidth]{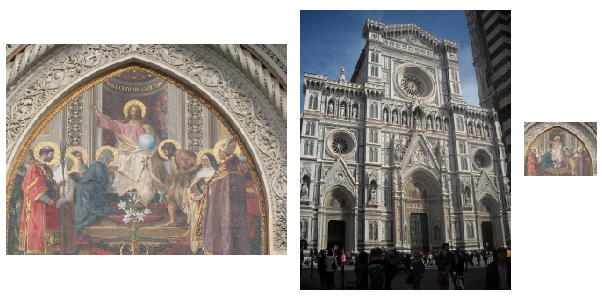} &

\includegraphics[width=0.21\textwidth]{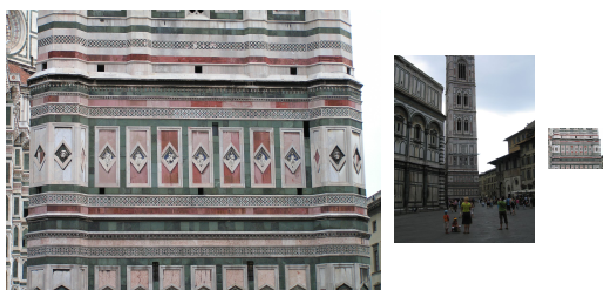} &

\includegraphics[width=0.21\textwidth]{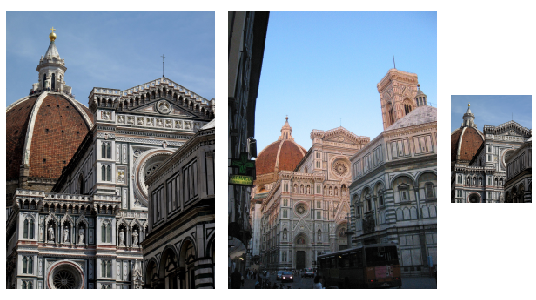} \\
\pred{65}{32} & \pred{80}{5} & \pred{71}{6} & \pred{70}{10}\\
\includegraphics[width=0.21\textwidth]{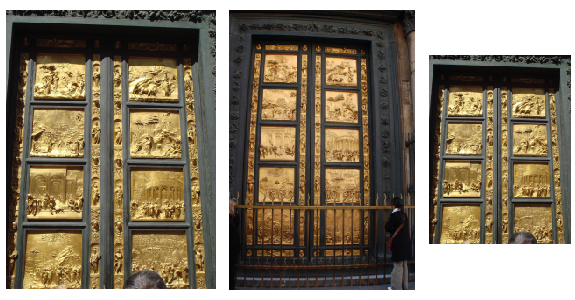} &

\includegraphics[width=0.21\textwidth]{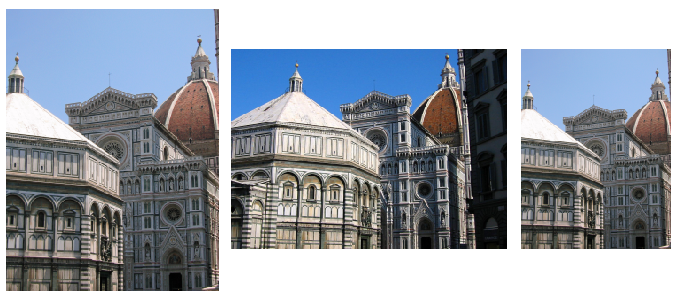} &

\includegraphics[width=0.2\textwidth]{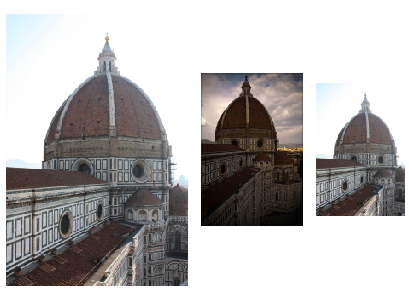} &

\includegraphics[width=0.21\textwidth]{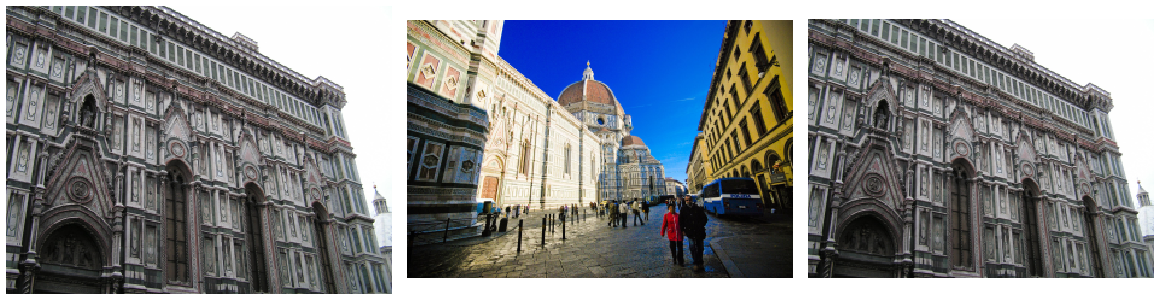} \\
\pred{94}{48} & \pred{88}{47} & \pred{76}{57} &\scriptsize{Failure:} \pred{43}{39} \\
\end{tabular}
\caption{Qualitative results of relative scale estimation on Florence test set. For each pair, the enclosure and concentration are calculated from which the relative estimated scaled can be derived. Based on that scale, the first image is resized and shown in the third position. If the relative scale is accurate, the content in the second and third images should match in size (number of pixels). The resized images are sometimes small, so the reader is encouraged to zoom into the images. The two numbers below each image pair show the estimated enclosure and concentration values. %
}
\label{fig:relative_scale}
\end{figure}

Figure~\ref{fig:relative_scale} shows qualitative examples of pairs of images from the test set, with predicted normalized box overlaps and estimated scale factor applied to one of the images.
We observed that this relative scale estimate is in general accurate if the images have zoom-in/zoom-out relationships. However, if the images are in crop-out/oblique-out relation to one-another, then the rescaling may not necessarily make matching easier, as the overlap is already quite small. However, we can detect if images are in such a relationship by looking at predicted box overlaps (see Section~\ref{sec:relationship}). %
The estimated scale factor can be applied to the images before local feature detection and description extraction, to improve relative pose accuracy for pairs of images with large scale differences. 

We simulate this scenario by retrieving $k=10$ and $k=30$ closest matches for each embedding, and solving for the pose. Additionally, for box embeddings, we do feature matching with and without pre-scaling of images according to the predicted overlaps. Figure~\ref{fig:boxplot} shows results on Notre-Dame and Big Ben scenes. 

Please see supplementary materials for further details.

\section{Conclusions}
We found surface overlap to be a superior measure compared to frustum overlap. We have shown that normalized surface overlap can be embedded in our new box embedding. The benefit is that we can now \emph{easily} compute interpretable relationships between pairs of images, without hurting localization. Further, this can help with pre-scaling of images for feature-extraction, and hierarchical organization of images. 

\paragraph{Limitations and Future work}
An obvious limitation is that we rely on expensive depth information for training. This could be addressed in two ways: either approximate the visual overlap with sparse 3D points and their co-visibility in two views, or train box embeddings with homography overlap of single images. Both of these approaches could also help with learning box embeddings that generalize across scenes, for image and object retrieval applications, as larger datasets could potentially be used for training.

\paragraph{Acknowledgements}
Thanks to Carl Toft for help with normal estimation, to Michael Firman for comments on paper drafts and to the anonymous reviewers for helpful feedback.

\bibliographystyle{splncs04}
\bibliography{main_bib}

\begin{thebibliography}{10}
\providecommand{\url}[1]{\texttt{#1}}
\providecommand{\urlprefix}{URL }
\providecommand{\doi}[1]{https://doi.org/#1}

\bibitem{arandjelovic2016netvlad}
Arandjelovic, R., Gronat, P., Torii, A., Pajdla, T., Sivic, J.: Netvlad: Cnn
  architecture for weakly supervised place recognition. In: CVPR (2016)

\bibitem{arandjelovic2013all}
Arandjelovic, R., Zisserman, A.: All about {VLAD}. In: CVPR (2013)

\bibitem{balntas2018relocnet}
Balntas, V., Li, S., Prisacariu, V.: {RelocNet}: Continuous metric learning
  relocalisation using neural nets. In: ECCV (2018)

\bibitem{balntas2016learning}
Balntas, V., Riba, E., Ponsa, D., Mikolajczyk, K.: Learning local feature
  descriptors with triplets and shallow convolutional neural networks. In:
  {BMVC} (2016)

\bibitem{bonin2008visual}
Bonin-Font, F., Ortiz, A., Oliver, G.: Visual navigation for mobile robots: A
  survey. Journal of intelligent and robotic systems  (2008)

\bibitem{brachmann2018learning}
Brachmann, E., Rother, C.: Learning less is more-{6D} camera localization via
  {3D} surface regression. In: CVPR (2018)

\bibitem{Buehler:2001}
Buehler, C., Bosse, M., McMillan, L., Gortler, S., Cohen, M.: Unstructured
  lumigraph rendering. In: Computer graphics and interactive techniques (2001)

\bibitem{Cakir-CVPR}
Cakir, F., He, K., Xia, X., Kulis, B., Sclaroff, S.: Deep metric learning to
  rank. In: CVPR (2019)

\bibitem{detone2016deep}
DeTone, D., Malisiewicz, T., Rabinovich, A.: Deep image homography estimation.
  arXiv preprint arXiv:1606.03798  (2016)

\bibitem{Dufournaud:2004}
Dufournaud, Y., Schmid, C., Horaud, R.: Image matching with scale adjustment.
  Computer Vision and Image Understanding  (2004)

\bibitem{erlik2017homography}
Erlik~Nowruzi, F., Laganiere, R., Japkowicz, N.: Homography estimation from
  image pairs with hierarchical convolutional networks. In: ICCVW (2017)

\bibitem{frahm2010building}
Frahm, J.M., Fite-Georgel, P., Gallup, D., Johnson, T., Raguram, R., Wu, C.,
  Jen, Y.H., Dunn, E., Clipp, B., Lazebnik, S., Pollefeys, M.: Building rome on
  a cloudless day. In: ECCV (2010)

\bibitem{GalvezTRO12}
G\'alvez-L\'opez, D., Tard\'os, J.D.: Bags of binary words for fast place
  recognition in image sequences. IEEE Transactions on Robotics  (2012)

\bibitem{gordo2016deep}
Gordo, A., Almaz{\'a}n, J., Revaud, J., Larlus, D.: Deep image retrieval:
  Learning global representations for image search. In: ECCV (2016)

\bibitem{guttman1984r}
Guttman, A.: R-trees: A dynamic index structure for spatial searching. In: ACM
  SIGMOD international conference on Management of data (1984)

\bibitem{hadsell2006dimensionality}
Hadsell, R., Chopra, S., LeCun, Y.: Dimensionality reduction by learning an
  invariant mapping. In: CVPR (2006)

\bibitem{hartley1997defense}
Hartley, R.I.: In defense of the eight-point algorithm. TPAMI  (1997)

\bibitem{he2016deep}
He, K., Zhang, X., Ren, S., Sun, J.: Deep residual learning for image
  recognition. In: CVPR (2016)

\bibitem{he2018local}
He, K., Lu, Y., Sclaroff, S.: Local descriptors optimized for average
  precision. In: CVPR (2018)

\bibitem{heinly2015reconstructing}
Heinly, J., Sch\"{o}nberger, J.L., Dunn, E., Frahm, J.M.: Reconstructing the
  world* in six days. In: CVPR (2015)

\bibitem{hermans2017defense}
Hermans, A., Beyer, L., Leibe, B.: In defense of the triplet loss for person
  re-identification. arXiv:1703.07737  (2017)

\bibitem{kendall2015posenet}
Kendall, A., Grimes, M., Cipolla, R.: {PoseNet}: A convolutional network for
  real-time {6-DOF} camera relocalization. In: ICCV (2015)

\bibitem{kulkarni2015deep}
Kulkarni, T.D., Whitney, W.F., Kohli, P., Tenenbaum, J.: Deep convolutional
  inverse graphics network. In: NeurIPS (2015)

\bibitem{laskar2017camera}
Laskar, Z., Melekhov, I., Kalia, S., Kannala, J.: Camera relocalization by
  computing pairwise relative poses using convolutional neural network. In:
  ICCV (2017)

\bibitem{le2020deep}
Le, H., Liu, F., Zhang, S., Agarwala, A.: Deep homography estimation for
  dynamic scenes. In: CVPR (2020)

\bibitem{li2018smoothing}
Li, X., Vilnis, L., Zhang, D., Boratko, M., McCallum, A.: Smoothing the
  geometry of probabilistic box embeddings. In: ICLR (2019)

\bibitem{MegaDepthLi18}
Li, Z., Snavely, N.: Megadepth: Learning single-view depth prediction from
  internet photos. In: CVPR (2018)

\bibitem{lowe2004distinctive}
Lowe, D.G.: Distinctive image features from scale-invariant keypoints. IJCV
  (2004)

\bibitem{mikulik2013browsing}
Mikulik, A., Chum, O., Matas, J.: Image retrieval for online browsing in large
  image collections. In: International Conference on Similarity Search and
  Applications (2013)

\bibitem{mikulik2014mining}
Mikul{\'\i}k, A., Radenovi{\'c}, F., Chum, O., Matas, J.: Efficient image
  detail mining. In: ACCV (2014)

\bibitem{NIPS2017-7068}
Mishchuk, A., Mishkin, D., Radenovic, F., Matas, J.: Working hard to know your
  neighbor's margins: Local descriptor learning loss. In: NeurIPS (2017)

\bibitem{mishkin2015mods}
Mishkin, D., Matas, J., Perdoch, M.: Mods: Fast and robust method for two-view
  matching. Computer Vision and Image Understanding  (2015)

\bibitem{orbslam2}
Mur-Artal, R., Tard\'os, J.D.: {ORB-SLAM2}: an open-source {SLAM} system for
  monocular, stereo and {RGB-D} cameras. IEEE Transactions on Robotics  (2017)

\bibitem{nguyen2018unsupervised}
Nguyen, T., Chen, S.W., Shivakumar, S.S., Taylor, C.J., Kumar, V.: Unsupervised
  deep homography: A fast and robust homography estimation model. IEEE Robotics
  and Automation Letters  (2018)

\bibitem{NIPS2017-7213}
Nickel, M., Kiela, D.: Poincar\'{e} embeddings for learning hierarchical
  representations. In: NeurIPS (2017)

\bibitem{perd2009efficient}
Perd'och, M., Chum, O., Matas, J.: Efficient representation of local geometry
  for large scale object retrieval. In: CVPR (2009)

\bibitem{perronnin2010large}
Perronnin, F., Liu, Y., S{\'a}nchez, J., Poirier, H.: Large-scale image
  retrieval with compressed {Fisher} vectors. In: CVPR (2010)

\bibitem{philbin2007object}
Philbin, J., Chum, O., Isard, M., Sivic, J., Zisserman, A.: Object retrieval
  with large vocabularies and fast spatial matching. In: CVPR (2007)

\bibitem{revaud2019r2d2}
Revaud, J., Weinzaepfel, P., De~Souza, C., Pion, N., Csurka, G., Cabon, Y.,
  Humenberger, M.: R2d2: Repeatable and reliable detector and descriptor. In:
  NeurIPS (2019)

\bibitem{sattler2011fast}
Sattler, T., Leibe, B., Kobbelt, L.: Fast image-based localization using direct
  {2D-to-3D} matching. In: ICCV (2011)

\bibitem{sattler2012improving}
Sattler, T., Leibe, B., Kobbelt, L.: Improving image-based localization by
  active correspondence search. In: ECCV (2012)

\bibitem{sattler2016efficient}
Sattler, T., Leibe, B., Kobbelt, L.: Efficient \& effective prioritized
  matching for large-scale image-based localization. TPAMI  (2016)

\bibitem{sattler2017large}
Sattler, T., Torii, A., Sivic, J., Pollefeys, M., Taira, H., Okutomi, M.,
  Pajdla, T.: Are large-scale {3D} models really necessary for accurate visual
  localization? In: CVPR (2017)

\bibitem{sattler2012image}
Sattler, T., Weyand, T., Leibe, B., Kobbelt, L.: Image retrieval for
  image-based localization revisited. In: BMVC (2012)

\bibitem{sattler2019understanding}
Sattler, T., Zhou, Q., Pollefeys, M., Leal-Taixe, L.: Understanding the
  limitations of cnn-based absolute camera pose regression. In: CVPR (2019)

\bibitem{schonberger2015paige}
Sch\"{o}nberger, J.L., Berg, A.C., Frahm, J.M.: Paige: pairwise image geometry
  encoding for improved efficiency in structure-from-motion. In: CVPR (2015)

\bibitem{schonberger2015single}
Sch\"{o}nberger, J.L., Radenovic, F., Chum, O., Frahm, J.M.: From single image
  query to detailed {3D} reconstruction. In: CVPR (2015)

\bibitem{schoenberger2016sfm}
Sch\"{o}nberger, J.L., Frahm, J.M.: Structure-from-motion revisited. In: CVPR
  (2016)

\bibitem{schroff2015facenet}
Schroff, F., Kalenichenko, D., Philbin, J.: Facenet: A unified embedding for
  face recognition and clustering. In: CVPR (2015)

\bibitem{shen2018matchable}
Shen, T., Luo, Z., Zhou, L., Zhang, R., Zhu, S., Fang, T., Quan, L.: Matchable
  image retrieval by learning from surface reconstruction. In: ACCV (2018)

\bibitem{shotton2013scene}
Shotton, J., Glocker, B., Zach, C., Izadi, S., Criminisi, A., Fitzgibbon, A.:
  Scene coordinate regression forests for camera relocalization in {RGB-D}
  images. In: CVPR (2013)

\bibitem{SGSS-siggraph08}
Snavely, N., Garg, R., Seitz, S.M., Szeliski, R.: Finding paths through the
  world's photos. ACM Transactions on Graphics  (2008)

\bibitem{snavely2006photo}
Snavely, N., Seitz, S.M., Szeliski, R.: Photo tourism: Exploring photo
  collections in {3D}. In: SIGGRAPH (2006)

\bibitem{sohn2012learning}
Sohn, K., Lee, H.: Learning invariant representations with local
  transformations. In: ICML (2012)

\bibitem{stewenius2012size}
Stew{\'e}nius, H., Gunderson, S.H., Pilet, J.: Size matters: exhaustive
  geometric verification for image retrieval. In: ECCV (2012)

\bibitem{subramanian2018new}
Subramanian, S., Chakrabarti, S.: New embedded representations and evaluation
  protocols for inferring transitive relations. In: SIGIR Conference on
  Research \& Development in Information Retrieval (2018)

\bibitem{tian2019sosnet}
Tian, Y., Yu, X., Fan, B., Wu, F., Heijnen, H., Balntas, V.: Sosnet: Second
  order similarity regularization for local descriptor learning. In: CVPR
  (2019)

\bibitem{torii201524}
Torii, A., Arandjelovic, R., Sivic, J., Okutomi, M., Pajdla, T.: 24/7 place
  recognition by view synthesis. In: CVPR (2015)

\bibitem{vendrov2015order}
Vendrov, I., Kiros, R., Fidler, S., Urtasun, R.: Order-embeddings of images and
  language. In: ICLR (2016)

\bibitem{vilnis2018probabilistic}
Vilnis, L., Li, X., Murty, S., McCallum, A.: Probabilistic embedding of
  knowledge graphs with box lattice measures. In: ACL (2018)

\bibitem{vilnis2014word}
Vilnis, L., McCallum, A.: Word representations via gaussian embedding. In: ICLR
  (2015)

\bibitem{iconoid1}
Weyand, T., Leibe, B.: Discovering favorite views of popular places with
  iconoid shift. In: ICCV (2011)

\bibitem{iconoid2}
Weyand, T., Leibe, B.: Discovering details and scene structure with
  hierarchical iconoid shift. In: ICCV (2013)

\bibitem{Witkin1983ScaleSpaceF}
Witkin, A.P.: Scale-space filtering. In: IJCAI (1983)

\bibitem{Worrall_2017_ICCV}
Worrall, D.E., Garbin, S.J., Turmukhambetov, D., Brostow, G.J.: Interpretable
  transformations with encoder-decoder networks. In: ICCV (2017)

\bibitem{zhou2017progressive}
Zhou, L., Zhu, S., Shen, T., Wang, J., Fang, T., Quan, L.: Progressive large
  scale-invariant image matching in scale space. In: ICCV (2017)

\end{thebibliography}
\end{document}